# Development and Validation of a Data Fusion Algorithm with Low-Cost Inertial Measurement Units to Analyze Shoulder Movements in Manual Workers

Marianne BOYER[1,2], Antoine FRASIE[2,3], Laurent BOUYER[2,3], Jean-Sébastien ROY[2,3], Isabelle POITRAS[2,3], and Alexandre CAMPEAU-LECOURS[1,2]

[1]Department of Mechanical Engineering, Université Laval, Canada, [2]Centre for Interdisciplinary Research in Rehabilitation and Social Integration, CIUSSS de la Capitale-Nationale, Canada, [3]Department of Rehabilitation, Université Laval, Canada

## ABSTRACT

Work-related upper extremity musculoskeletal disorders (WRUED) are a major problem in modern societies as they affect the quality of life of workers and lead to absenteeism and productivity loss. According to studies performed in North America and Western Europe, their prevalence has increased in the last few decades. This challenge calls for improvements in prevention methods. One avenue is through the development of wearable sensor systems to analyze worker's movements and provide feedback to workers and/or clinicians. Such systems could decrease the physical work demands and ultimately prevent musculoskeletal disorders. This paper presents the development and validation of a data fusion algorithm for inertial measurement units to analyze worker's arm elevation. The algorithm was implemented on two commercial sensor systems (Actigraph GT9X and LSM9DS1) and results were compared with the data fusion results from a validated commercial sensor (XSens MVN system). Cross-correlation analyses [r], root-mean-square error (RMSE) and average absolute error of estimate were used to establish the construct validity of the algorithm. Five subjects each performed ten different arm elevation tasks. The results show that the algorithm is valid to evaluate shoulder movements with high correlations between the results of the two different sensors and the commercial sensor (0.900-0.998) and relatively low RMSE value for the ten tasks (1.66-11.24°). The proposed data fusion algorithm could thus be used to estimate arm elevation.

## INTRODUCTION

Work-related shoulder injuries impact negatively the quality of life of workers and lead to problems of absenteeism and loss of productivity [1]. Despite prevention efforts, prevalence of these injuries is increasing [2]. Given the importance of this issue, workplace interventions must be improved. Studies have shown that multiple factors such as posture, force, amount of repeated movements and range of motion should be taken into account in the study of physical effort causing musculoskeletal injuries [3].

The three principal methods to analyse human motion are: 1) qualitative assessment (observation with naked eye by an evaluator), 2) quantitative, fixed camera-based motion capture systems (mainly lab-based), and 3) direct, in-the-field quantitative methods using wearable sensors. Despite its accuracy in the laboratory, method #2 is too cumbersome to be easily usable in real work environments. On the contrary, wearable sensors have the benefit of being usable in any environment, from the laboratory to the workplace. For example, inertial measurement units (IMUs) have been shown to be valid tools to assess shoulder movements during simple arm elevations as well as during complex lifting tasks [4].

The main challenge with commercial sensors is that they were developed for research or sports applications, but not necessarily for clinical rehabilitation purposes. As such, they provide neither appropriate interfaces nor adequate feedback for clinical applications. Furthermore, their architecture is closed which prevents the addition of non-proprietary hardware (e.g., vibration motor to alert the user) or to adjust the algorithms for rehabilitation needs. Regarding non-commercial solutions, many open-source algorithms can be used to find IMUs orientation from raw data and can be implemented on custom devices [5,6]. However, these algorithms differ in quality, and have not been validated for neither general nor rehabilitation purposes.

## OJECTIVES

The long-term aim of this project is therefore to develop a low-cost wearable device using inertial measurement units (IMUs) able to analyze shoulder movement and provide feedback to clinicians and workers (both offline and in real time), to reduce the risk of musculoskeletal injuries. The specific objectives of this paper are 1) to develop an IMU data fusion algorithm to estimate the shoulder elevation and 2) validate the latter, when implemented on two different low-cost sensors (Actigraph GT9X and LSM9DS1), by comparing it to the elevation angle obtained with the data fusion algorithm of a validated commercial sensor system (XSens MVN).

**METHODS : DATA FUSION ALGORITHM**

Arm orientation can be described by three angles: plane of elevation, segment elevation and internal/external rotation. In this paper, we are mainly interested by segment elevation as it is the principal indicator related to the development of musculoskeletal disorders. Furthermore, while arm elevation can be obtained solely from IMU's accelerometer and gyroscope data, the plane of elevation and the rotation require additional information from a magnetometer, which is known to lack robustness due to its sensitivity to local magnetic disturbances. The proposed algorithm's inputs are the IMU 3-axis accelerations and angular velocities while the output is arm elevation. At first, a static calibration is performed where the user's arm remains still. The sensor's orientation at rest is thus obtained by using the acceleration ratio between each axis[1]. This static calibration phase is also used to calibrate the gyroscopes, which are known to be naturally drifting. To this end, the mean velocity is acquired for all three axes during the calibration and subtracted from the gyroscope's data at each further iteration. Then, the raw data (from the accelerometer and gyroscope) is acquired and processed to estimate arm elevation at each time step. The process is shown in Figure 1.

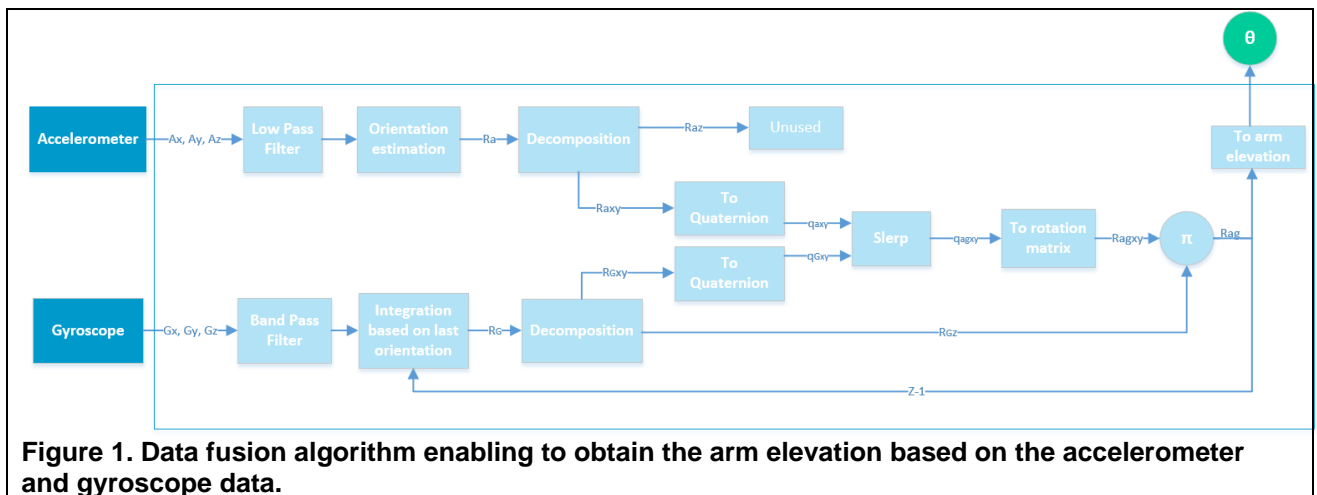

**Figure 1. Data fusion algorithm enabling to obtain the arm elevation based on the accelerometer and gyroscope data.**

Accelerations are passed trough a low-pass filter (order 1 at 50Hz) to reduce high-frequency noise, while angular velocities from the gyroscope are passed trough a band-pass filter (order 1 between 0.002Hz and 50Hz) to reduce high-frequency noise and minimize drifting. The ratios between each component (X,Y,Z) of the acceleration are used to estimate the sensor's rotation matrix at each time step[1]. The sensor's rotation matrix is also independently updated from the last know orientation by integrating the resultant angular velocity vector obtained from the three axis of the gyroscopes. As the accelerometer does not provide information on sensor orientation perpendicular to gravity (around the world Z axis), both orientation matrices (obtained independently from the accelerometer and from the gyroscope) are decomposed into two matrices, namely one for the world XY rotations and another for rotations along the world Z axis, for a total of four matrices. Both XY rotation matrices are then transformed into quaternions to avoid representation singularities. From that representation, both quaternions are combined using spherical linear interpolation (Slerp) and transformed into a rotation matrix. The latter rotation matrix is then multiplied with the world Z axis rotation matrix obtained from the gyroscope's update to obtain the final rotation matrix. Finally, arm elevation is found by transforming the rotation matrix into tilt and torsion angles [7].

**METHODS : VALIDATION**

Five healthy adults with no self-reported musculoskeletal conditions (i.e. pain or movement limitations) took part in one testing session (1 woman and 4 men, right-handed, 23-44 years old). Right shoulder movements were recorded simultaneously by three IMUs (XSens MVN, ActiGraph GT9X and LSM9DS1) positioned on the lateral aspect of the right arm, at its distal end (Figure 2a). The LSM9DS1 data was acquired thanks to an ATMEL ARM Cortex m7 microcontroller through i2c communication. The IMUs were attached with hook and loop straps around the arms and positioned on a Lycra suit for the trunk, in accordance with the sensors' configuration recommended by Xsens. For the Xsens MVN IMUs, raw data are internally obtained at 1000 Hz and segmentelevation are provided at 100 Hz with *MT Manager software version 4.6*. For the ActiGraph GT9X and LSM9DS1 sensors, the raw data were obtained respectively at 100 Hz and 500 Hz, and the segment elevation was obtained with the proposed data fusion algorithm.

---

[1] https://arduino.onepamop.com/wp-content/uploads/2016/03/AN3461.pdf

The validation session began with an anatomical pose (participant standing straight and looking forward, arms along body side, palms facing the thighs) to calibrate the three systems. The protocol consisted of 10 tasks, which were each performed five times by participants: 1) shoulder flexion at 1Hz, 2) shoulder external rotation at 90° elbow flexion (1 Hz), 3) shoulder flexion at 3Hz, 4) shoulder abduction at 1Hz, 5) shoulder external rotation at 90° abduction (1 Hz), 6) shoulder abduction at 3 Hz, 7) trunk flexion with static arm elevation (Figure 2b), 8-9) five "Z" movements on the frame of a mirror (Figure 2c) (clockwise and counterclockwise), 10) nine ball throws at 90° shoulder abduction (target distance of 2.85 m; Figure 2d). Movement frequency was maintained using a metronome.

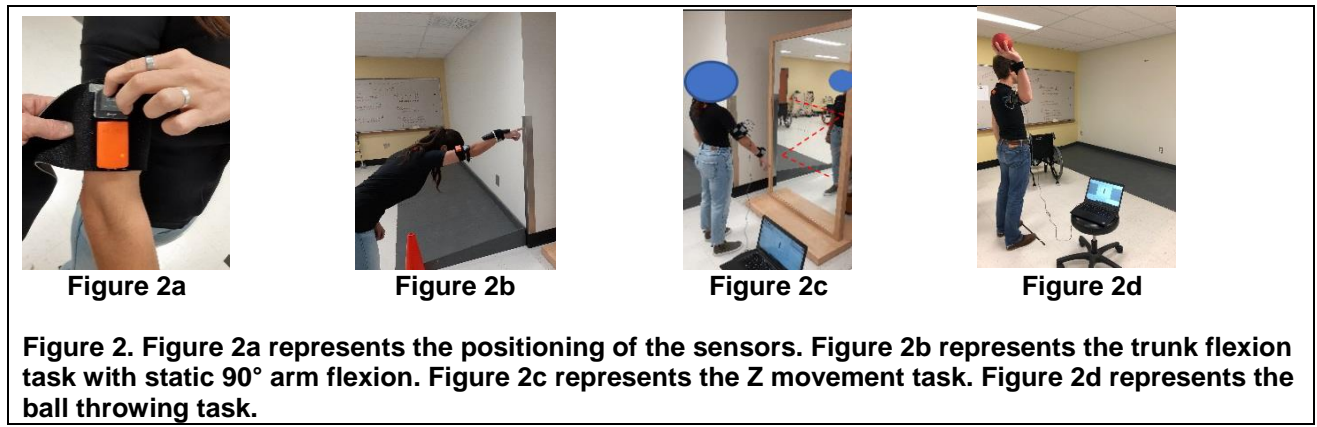

**Figure 2a**  **Figure 2b**  **Figure 2c**  **Figure 2d**

**Figure 2. Figure 2a represents the positioning of the sensors. Figure 2b represents the trunk flexion task with static 90° arm flexion. Figure 2c represents the Z movement task. Figure 2d represents the ball throwing task.**

Statistical analysis. Cross-correlation analyses [r] were performed on arm elevation for XSens MVN vs LSM9DS1 and XSens MVN vs GT9X data to establish the convergent construct validity. Arm elevation errors (XSens MVN vs LSM9DS1 and XSens MVN vs GT9X) were compared for each task using root-mean-square error calculation (RMSE) and average absolute error of estimate.

## RESULTS

Figure 3 presents arm elevation for tasks 1 to 10 for a representative participant. Table 1 presents cross-correlation analyses (tasks 2,5,7 are not presented as arm elevation remained quasi-static), RMSE and average absolute error of estimate. The correlation coefficients (r) were excellent for GT9X (0.900-0.998) and LSM9DS1 (0.900-0.998) for tasks 1 to 9 but were lower with GT9X (0.894) for task 10. LSM9DS1 showed lower RMSE and average absolute error of estimate (RMSE = 1.66-11.24°; absolute error of estimate = 1.23-7.50°).

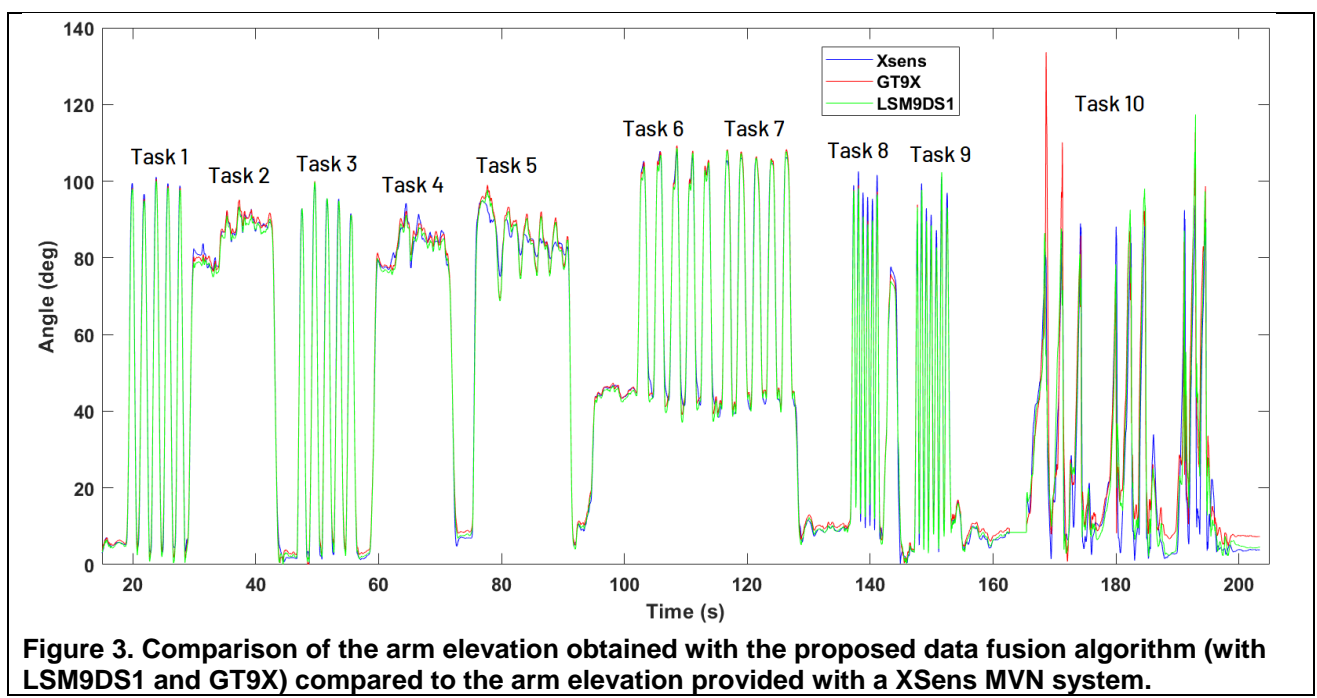

**Figure 3. Comparison of the arm elevation obtained with the proposed data fusion algorithm (with LSM9DS1 and GT9X) compared to the arm elevation provided with a XSens MVN system.**

**Table 1 - Correlation coefficient, root-mean-square error, average absolute error of estimate.**

| Task | r_GT9X (Mean[SD]) | r_LSM (Mean[SD]) | RMSE_GT9X (Mean[SD])(°) | RMSE_LSM (Mean[SD])(°) | Averrage Error Estimate_GT9X (Mean[SD])(°) | Averrage Error Estimate_LSM (Mean[SD])(°) |
|---|---|---|---|---|---|---|
| 1- Flexion (1 Hz) | 0,986[0,027] | 0,998[0,001] | 4,47[2,67] | 2,91[0,54] | 3,61[1,68] | 2,47[0,46] |
| 2- Ext. rotation at 90° flexion (1 Hz) | NA | NA | 2,20[1,79] | 1,66[0,62] | 1,86[1,79] | 1,23[0,53] |
| 3- Flexion (3 Hz) | 0,964[0,042] | 0,973[0,032] | 9,95[4,27] | 9,25[4,65] | 7,73[3,39] | 7,18[3,61] |
| 4- Abduction (1 Hz) | 0,998[0,001] | 0,998[0,000] | 3,05[1,11] | 2,37[0,55] | 2,48[0,93] | 1,75[0,43] |
| 5- Ext. rotation at 90° abduct (1 Hz) | NA | NA | 2,13[1,49] | 1,87[0,70] | 1,87[1,54] | 1,59[0,69] |
| 6- Abduction (3 Hz) | 0,982[0,011] | 0,988[0,009] | 7,44[1,36] | 6,23[1,68] | 5,47[0,88] | 4,50[1,06] |
| 7- Trunk flexion, static arm flexion | NA | NA | 3,74[2,12] | 2,91[1,03] | 3,18[2,05] | 2,33[0,86] |
| 8- "Z" movements clockwise | 0,996[0,003] | 0,993[0,005] | 3,37[1,51] | 3,62[1,30] | 2,78[1,26] | 2,93[1,14] |
| 9-"Z" movements counterclockwise | 0,995[0,002] | 0,995[0,003] | 3,88[2,27] | 3,70[1,33] | 3,25[2,13] | 3,12[1,18] |
| 10- Ball throws | 0,894[0,052] | 0,900[0,045] | 11,25[1,73] | 11,24[1,85] | 8,00[1,39] | 7,50[1,03] |

**DISCUSSION AND CONCLUSION**

The objectives of this project were to: 1) develop an IMU data fusion algorithm to estimate shoulder elevation and 2) validate it, when implemented on two different low-cost sensors (Actigraph GT9X and LSM9DS1), by comparing it to the elevation angle obtained with the data fusion algorithm of a validated commercial sensor system (XSens MVN). The results show a high correlation (r > 0.90) for all tasks and a mean RMSE error below 4.6° (1.66-11.24°) for LSM. The proposed data fusion algorithm is thus valid to estimate arm elevation. This algorithm yields better results for slower (tasks 1,2,4,5,7,8,9 with a mean RMSE of 2.72° for LSM) than for faster movements (tasks 3,6,10 with a mean RMSE of 8.9° for the LSM). The results obtained with the LSM9DS1 were better than Actigraph GT9X. While the same data fusion algorithm was used with both sensors, the higher sampling rate of LSM9DS1 (500Hz vs 100Hz) could explain these results.

The long-term objective of this work is to develop a low-cost wearable device using IMUs to analyze shoulder movements and provide feedback to clinicians and workers to reduce the risk of musculoskeletal injuries. Future work will consist in validating the system in a workplace environment, to miniaturize the system and to provide a meaningful data report to clinicians using the arm elevation data obtained throughout a day.

**ACKNOWLEDGEMENTS**

This work was supported by the Canada First Research Excellence Fund Sentinel North Strategy at Laval University, the Canadian MSK Rehab Research Network and Dr. Campeau-Lecours's startup funds at Cirris.